\documentclass[conference]{ieeeconf}

\usepackage{graphicx}  
\usepackage[boxed]{algorithm2e}
\usepackage{amsmath,amssymb}
\usepackage{ctable}
\usepackage{hhline}
\usepackage{xcolor}
\usepackage{colortbl}
\usepackage{multirow}
\usepackage{subfigure}
\usepackage{multirow}

\begin{document}
	%
	\title{A Generative Modeling Approach to Limited Channel ECG Classification}
	\author{ Deepta Rajan$^*$ and Jayaraman J. Thiagarajan$^{\dagger}$ \\
		$^*$IBM Almaden Research Center, 650 Harry Road, San Jose, CA. \\
		$^{\dagger}$Lawrence Livermore National Laboratory, 7000 E Avenue, Livermore, CA 94550.
	}
	\maketitle
	\begin{abstract}
Processing temporal sequences is central to a variety of applications in health care, and in particular multi-channel Electrocardiogram (ECG) is a highly prevalent diagnostic modality that relies on robust sequence modeling. While  Recurrent Neural Networks (RNNs) have led to significant advances in automated diagnosis with time-series data, they perform poorly when models are trained using a limited set of channels. A crucial limitation of existing solutions is that they rely solely on discriminative models, which tend to generalize poorly in such scenarios. In order to combat this limitation, we develop a generative modeling approach to limited channel ECG classification. This approach first uses a \textit{Seq2Seq} model to implicitly generate the missing channel information, and then uses the latent representation to perform the actual supervisory task. This decoupling enables the use of unsupervised data and also provides highly robust metric spaces for subsequent discriminative learning. Our experiments with the Physionet dataset clearly evidence the effectiveness of our approach over standard RNNs in disease prediction.
	\end{abstract}

\section{Introduction}

With the unprecedented success of machine learning in solving challenging problems across multiple domains, there is increasing interest in leveraging state-of-the art techniques to applications in health care. The community-wide efforts for creating large-scale benchmark repositories, such as MIMIC-III and Physionet CinC challenge \cite{clifford2017af}, have accelerated machine learning research in health care. Furthermore, with increased adoption of automated systems for disease diagnosis, there is a huge opportunity for building robust data-driven solutions that can alleviate pain-points within clinical workflows. Broadly, careful modeling of health care data requires tackling inherent challenges including multi-variate measurements, long-range temporal dependencies, and missing information in order to make precise predictions. Despite the success of hand-engineered features in clinical models, more recently, regularized representation learning techniques, such as sparse and deep learning, have been more effective. A thorough experimental study on UCR time-series datasets revealed that simple deep learning architectures using 1-D Convolutional Neural Networks (CNNs) can easily outperform traditional task-specific models built on hand-engineered features \cite{wang2017time}. More recently, Recurrent Neural Networks (RNN) based on Long Short Term Memory (LSTM) units have become the de-facto solution for clinical time-series analysis. Choi et. al \cite{choi2016doctor} demonstrated the effectiveness of LSTMs to deal with clinical sequence classification. The state-of-the-art Intensive Care Unit (ICU) sequence modeling architectures are based on deep attention models that solve multiple prediction tasks jointly \cite{song2017attend}. 

\subsection{Problem Statement} In this work, we investigate classification of multi-channel Electrocardiograms (ECG) measurements, one of the most common diagnostic modalities. The overall goal is to learn features from the multi-variate sequences that can potentially help diagnose heart conditions such as \textit{Myocardial Infarction, Branch Bundle Blocks, Cardiomyopathy}, and several others \cite{PhysioNet}. Generally speaking, to accurately identify if an ECG depicts a normal sinus rhythm or an abnormal rhythm requires detection of various wave segments (P-wave, QRS-complex, T-wave) and understanding their complex morphological relationships over time. This motivated the design of classical signal processing approaches such as the Pan-Tomkins algorithm \cite{pan1985real} and wavelet analysis \cite{martinez2004wavelet}. However, this task remains challenging due to its episodic nature, inherent measurement noise, inter-patient variability of wave patterns, and ambiguity of labels etc. Consequently, neural network based solutions have been developed for this problem, in particular for detecting \textit{Myocardial Infarction}. Similar to \cite{acharya2017application, reasat2017detection}, in this paper, we consider an additional challenge that it is necessary to perform predictions using a limited channel configuration at test time. We show that, RNNs are plagued by overfitting, and they produce inferior predictions under the limited channel setting.

\begin{figure*}[t]
	\centering
	\includegraphics[width = 1\linewidth]{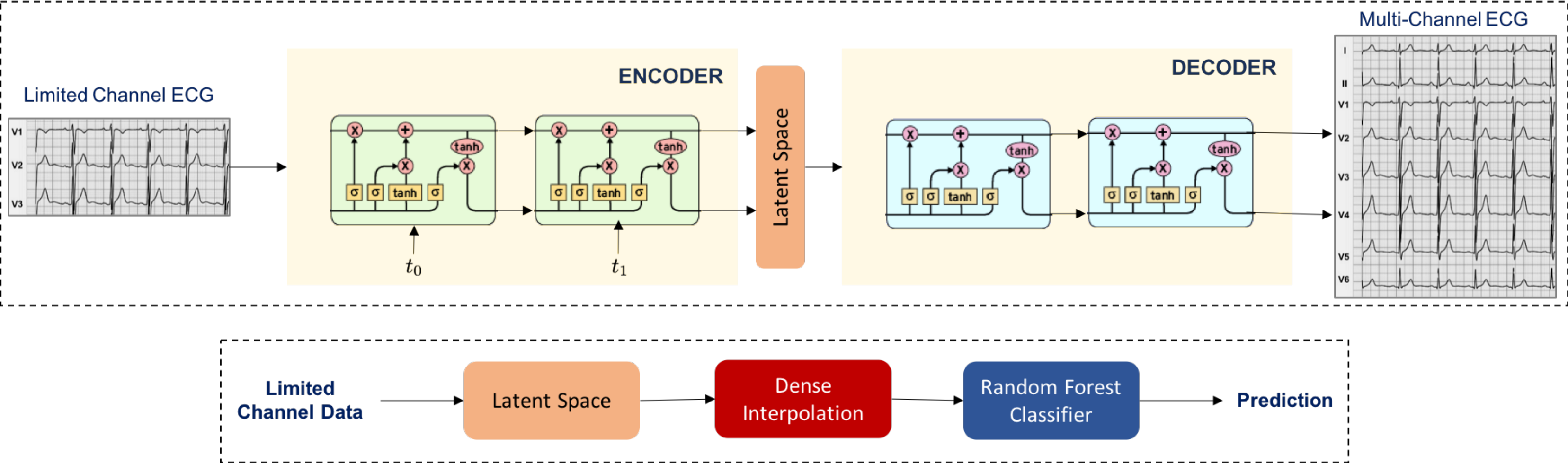}
	\caption{Proposed approach for limited channel ECG classification. In the first stage, we adopt a unsupervised, generative modeling approach to construct the latent space, while in the second stage we aggregate the temporal features from the latent space using dense interpolation and use them to train a random forest classifier. }
	\label{fig:overview}
\end{figure*}

\subsection{Proposed Work} Inspired by the recent surge in the success of generative models for image \cite{li2017max} and text classification \cite{yogatama2017generative}, we propose a generative modeling approach to limited channel ECG classification. Unsupervised generative models, e.g. variational auto-encoders \cite{kingma2013auto} and Seq2Seq models \cite{sutskever2014sequence}, enable the inference of latent features that can effectively describe the complex distribution of data. In our approach, we utilize \textit{Seq2Seq} modeling to construct latent spaces that can predict the entire multi-channel (12-channel in our case) ECG measurements using only the limited channel data. The resulting latent feature representations implicitly exploit information from the missing channels and can generalize better for limited channel measurements at test time. In the classifier design stage, the latent representations are aggregated into feature vectors using a dense interpolation scheme \cite{song2017attend} and are subsequently used to train a simple random forest classifier. Our experiments with the PhysioNet PTB dataset \cite{PhysioNet} show that, the proposed generative approach is significantly superior to the conventional discriminative architecture, even in the 12-channel case, while remaining robust to limited channel conditions. Given the critical need to make accurate predictions from limited measurements (single-channel ECG) recorded by mobile ECG monitors in practical scenarios \cite{rajpurkar2017cardiologist}, \cite{clifford2017af}, our generative modeling based approach can generalize well to different datasets and measurement configurations.





\section{Proposed Approach}
Discriminative models have been commonly employed for a variety of multivariate clinical time-series classification tasks. In particular, sequence modeling techniques based on Recurrent Neural Networks (RNNs) have achieved state-of-the-art results in multi-channel ECG classification. Despite their widespread use, we show that discriminative architectures, which jointly learn the latent features and the classifier, can be non-robust when dealing with datasets characterized by imbalanced class distributions. In such scenarios, the models are often plagued by overfitting, and can hence produce highly unreliable predictions. Furthermore, in cases where we consider only a subset of channels for the actual prediction, RNNs based on LSTM units (Long-Short Term Memory) perform very poorly, thus motivating the need to design a robust prediction pipeline. In this section, we describe a novel, generative modeling approach to multi-variate sequence classification, which can handle both imbalanced class distributions and limited channel scenarios. 

\subsection{Overview}
Figure \ref{fig:overview} illustrates an overview of the proposed approach for limited channel ECG classification. The key idea of our approach is to employ a generative architecture to model the multi-variate sequences, prior to learning the classifier. Broadly referred to as \textit{classification by synthesis}, such an approach has been adopted in image classification, particularly to deal with out-of-distribution samples at test time \cite{wang2017safer}. In this paper, we argue that using a generative approach can produce models that are highly robust to using partial measurements for prediction. In a nutshell, the proposed approach is comprised of two stages: (i) unsupervised generative modeling stage where we utilize a Seq2Seq architecture to construct a latent space that is the most effective for predicting multi-channel ECG data using only a partial set of channels; (ii) supervised modeling stage that builds a random forest classifier using latent features from Stage 1. Note that, in order to aggregate features from different time steps, while preserving partial order, we propose to employ a dense interpolation strategy, similar to \cite{song2017attend}. At the testing stage, the predictions are made solely based on the partial set of channels that was used to train the model.

\subsection{Stage 1: Generative Models for Limited Channel Data}
As shown in Figure \ref{fig:overview}, we assume access to the entire multi-channel ECG measurements for the training data, and for a given limited channel configuration, we enable predictions using only the subset of channels at test time. Denoting the multi-variate sequence dataset as $X \in \mathbb{R}^{N \times T \times K}$, where $N$ denotes the number of training samples, $T$ denotes the number of time-steps in each measurements and $K$ indicates the total number of channels. The channel configuration is typically determined based on the disease to be diagnosed, for example, the channels \{\textit{II, III, aVF}\} are known to be essential for detecting \textit{Myocardial Infarction} \cite{reasat2017detection}. Denoting the set of limited channels by $\mathcal{C}$, whose cardinality $\hat{K} \ll K$, we extract the matrix $\hat{X} \in \mathbb{R}^{N \times T \times \hat{K}}$. In order to perform implicit completion of the missing data, we propose to build a generative model that attempts to recover $X$ using $\hat{X}$. In this process, it infers a latent space that defines an effective metric to compare different samples.

More specifically, we build an encoder-decoder architecture, commonly referred as Seq2Seq \cite{sutskever2014sequence}, with an optional attention mechanism. Though originally developed for machine translation, they are applicable to more general sequence to sequence transformation tasks. The architecture is comprised of two RNNs (based on LSTM), one each for encoder and decoder. The encoder transforms an input sequence from $\hat{X}$ into a fixed length vector, either from the last time step of the sequence or by concatenating hidden representations from all time steps. The decoder then predicts the output sequence, in our case $X$, using the encoder output.  Optionally, the decoder can also attend to a certain part of the encoder states through an attention mechanism. The attention mechanism often uses both content from the encoder states, and also context from the sequence generated so far at the decoder. Our RNNs are designed using Long Short Term Memory units, which are capable of learning long-term dependencies. Each LSTM cell is comprised of the following operations, implemented using fully connected networks:
\begin{align}
\nonumber &\text{(input gate):\quad} i^{(t)} = \sigma(W^{(i)} x^{(t)} + U^{(i)} h^{(t - 1)}) \\
\nonumber&\text{(forget gate):\quad} f^{(t)} = \sigma(W^{(f)} x^{(t)} + U^{(f)} h^{(t - 1)})
\\
&\text{(output gate):\quad} o^{(t)} = \sigma(W^{(o)} x^{(t)} + U^{(i)} h^{(t - 1)})
\\
\nonumber&\text{(new cell):\quad} \bar{c}^{(t)} = tanh(W^{(c)} x^{(t)} + U^{(c)} h^{(t - 1)})
\\
\nonumber&\text{(final cell):\quad} i^{(t)} = f^{(t)} o \text{ } \bar{c}^{(t -1)} + i^{(t)} o \text{ } \bar{c}^{(t)}.
\end{align}The LSTM has the ability to remove or add information to the cell state, carefully regulated by structures referred to as gates. While the \textit{forget gate} updates the cell states by determining which information to ignore based on context, the \textit{input gate} determines which information needs to be updated to the cell state based on the previous hidden state and content at the current time step. Finally the \textit{output gate} produces a filtered version of the cell state, based on the context and previous hidden state. The generative model is trained with an $\ell_2$ loss at the decoder output. Note that, our architecture attempts to reconstruct the input channels as well as predict the missing channel measurements.

\subsection{Stage 2: Classifier Design}
We now design a classifier stage that exploits the latent space from the generative model trained for missing channel prediction. Interestingly, compared to discriminative models, this approach utilizes additional channel information from the training stage and builds a more effective metric for the whole data space instead of discriminating the normal/abnormal classes. Furthermore, since the first stage is unsupervised, we can use even unlabeled data to construct a more robust latent space.

\subsubsection{Dense Interpolation Embedding}
For given limited channel sequences, $\hat{X}$, the encoder returns the latent features $Z \in \mathbb{R}^{N \times T \times h}$, where $h$ is the number of hidden dimensions in the Seq2Seq model. The simplest approach to obtain an aggregated representation for this feature, while preserving order, is to simply concatenate embeddings at every time step. However, in our case, this can lead to a very high-dimensional, ``cursed'' representation which is not suitable for learning and inference. Consequently, we propose to utilize a  dense interpolation strategy similar to \cite{song2017attend}. Denoting the hidden representation at time $t$ as $z_t\in \mathbb{R}^{h}$, the interpolated embedding vector will have dimension $h \times M$, where $M$ is the \textit{dense interpolation factor}. Note that when $M=T$, it reduces to the concatenation case. The main idea of this scheme is to determine weights $w$, denoting the contribution of $z_t$ to the position $m$ of the final vector representation $s$.

\subsubsection{Classifier}
Using the embeddings from dense interpolation, $S \in \mathbb{R}^{N \times M.h}$, we build a random forest classifier to predict the labels $\mathcal{Y} \subset \{-1, +1\}$. At test time, each limited channel sequence is passed through the encoder of the generative model to obtain the latent representation and subsequently processed using the dense interpolation strategy and the random forest classifier.

\section{Experiments}
In this section, we describe the dataset used in the experiments, and evaluate our proposed generative modeling framework on the task of ECG classification. In addition, we discuss the evaluation metrics, and the choices for the hyperparameters. Finally, we elaborate on the disease-based limited channel configurations selected for experimentation and highlight the comparative performance of standard RNNs against the proposed approach.

The Physionet PTBDB \cite{PhysioNet} is comprised of ECGs collected from healthy volunteers as well as patients with a wide range of heart diseases. It has a total of $549$ records from $290$ subjects, with each record containing the standard 12-leads along with 3 Frank lead ECGs (vx, vy, vz). A single raw ECG record is about 30 seconds in duration with measurements sampled at 1000 Hz. As part of data-preprocessing, we normalized the data, defined an ECG frame to be a 5-seconds long, with a 2-second overlap between frames and finally down-sampled them to $500$ time-steps.

\begin{table}[]
	\renewcommand*{\arraystretch}{1.2}
	\centering
	\caption{AUROC scores of normal/abnormal binary classification task for different channel configurations. We report the performance of standard RNNs and the proposed approach. In each case, the best performance is marked in bold.}
	\label{table:perf}
	\begin{tabular}{|c|c|c|}
		\hline
		\rowcolor[HTML]{C0C0C0} 
		\textbf{Channels} & \textbf{Standard RNN}        & \multicolumn{1}{l|}{\cellcolor[HTML]{C0C0C0}\quad \textbf{Proposed} \quad} \\ \hline
		 \hline
		\rowcolor[HTML]{FFFFFF} 
		V1 to V6          & 0.78                         & \textbf{0.83}                                                  \\ \hline
		V1, V2, V3        & 0.60                        & \textbf{0.69}                                                  \\ \hline
		\rowcolor[HTML]{FFFFFF} 
		II, III, aVF      & 0.56                        & \textbf{0.60}                                                  \\ \hline
		II, III, V3       & 0.55                          & \textbf{0.69}                                                  \\ \hline
		\rowcolor[HTML]{FFFFFF} 
		V1, V6            & \cellcolor[HTML]{FFFFFF}0.53 & \cellcolor[HTML]{FFFFFF}\textbf{0.74}                          \\ \hline
		II, V1            & 0.52                         & \textbf{0.59}                                                  \\ \hline
		II                & 0.51                          & \textbf{0.62}                                                  \\ \hline
	\end{tabular}
\end{table}

\begin{table}[t]
	\centering
	\renewcommand*{\arraystretch}{1.2}
	\caption{Performance of disease-specific models trained using appropriate channel subsets with the proposed approach. }
	\label{table:disease}
	\begin{tabular}{|c|c|l|l|l|}
		\hline
		\rowcolor[HTML]{C0C0C0} 
		\textbf{Disease}                           & \multicolumn{1}{l|}{\cellcolor[HTML]{C0C0C0}{\color[HTML]{333333} \textbf{Channels}}} & \textbf{Acc.$\%$}         & \textbf{Sens.$\%$}      & \textbf{Spec.$\%$}      \\ \hline
		Myocardial Infarction & V1, V2, V3                                                                                         & \multicolumn{1}{c|}{86} & \multicolumn{1}{c|}{96} & \multicolumn{1}{c|}{84} \\ \hline
		\multicolumn{1}{|l|}{Bundle Branch Block} & V1, V6                                                                                             & \multicolumn{1}{c|}{94}                          &    \multicolumn{1}{c|}{97}                      & \multicolumn{1}{c|}{99}                       \\ \hline
	\end{tabular}
\end{table}

\subsection{Setup and Evaluation Metrics}
\subsubsection{Task I: Normal/Abnormal Classification}
Although PTBDB is a multi-class dataset, containing $8$ distinct abnormal conditions, a miscellaneous class and a healthy control group, in this task, we formulate the problem into a two-class problem of separating normal and abnormal ECGs by grouping all disease classes into a single abnormal class. This makes the problem highly ill-posed, challenging a classifier to ignore the large variations inside the abnormal class while discriminating from the normal class. Problems using clinical data are commonly solved with such a normal versus abnormal formulation, with an optional second stage classification to determine the actual type of abnormality. However, in practice, this formulation makes it very challenging for discriminative models (e.g. standard RNNs). In contrast, our approach exploits additional data at the training stage in the form of missing channel information and consequently produces more reliable latent spaces. As we will show in our results, this two stage approach is significantly superior to standard RNNs, even when we reduce the number of channels drastically ($\hat{K} \leq 3$). In this task, we used $5517$ frames for training and $2718$ frames for testing. For this task, we use a popular summary statistic metric - micro-averaged AUROC, which computes the area under the ROC curve for both classes together.

\subsubsection{Task II: Disease-specific Prediction}
In this task, we consider a subset of the data corresponding to specific abnormalities, \textit{Myocardial Infarction} and \textit{Bundle Branch Block}, and build a predictive model to detect that condition. For both cases, we used channel subsets that are commonly used in clinical settings for prediction of those diseases (Table \ref{table:disease}). In the case of \textit{Myocardial Infarction}, we used $4502$ frames for training and $2218$ for testing. Whereas, for \textit{Branch Bundle Blocks}, we used $974$ frames for training and $481$ for testing. Following common practice, we use three metrics: $\%$Accuracy, $\%$Sensitivity and $\%$Specificity.

For both standard RNN and the proposed approach, we considered $2$-layer LSTM networks with $h = 64$. The dense interpolation factor $M$ was set at $16$, thus producing feature vectors of dimension $1024$. All models were trained using Adam optimizer with learning rate $0.001$.  

\subsection{Results}
Table \ref{table:perf} reports the results for \textit{Task I} obtained using standard RNNs and the proposed approach, under different channel configuration settings. As discussed earlier, when the entire 12-lead data is not available at test time, it is important for models to generalize to limited channel scenarios. As it can be observed from the results, the proposed approach is significantly better than standard RNNs for all cases, with the AUROC metric indicating its reliability in predicting both normal and abnormal class samples. For example, in the case of using channels $V1$ to $V6$, the proposed approach achieves an AUROC of $0.83$, which is superior than $0.78$ achieved by standard RNNs with similar number of parameters. Interestingly, even in the case of a single channel, the proposed approach is more effective in dealing with the imbalance in label distribution, providing highly robust predictions in comparison to discriminative models.

The results for disease-specific classification models are shown in Table \ref{table:disease}. Though not reported, we found similar improvements over standard RNNs as observed in the previous task. For example, in the case of  \textit{Myocardial Infarction} detection, the proposed approach achieves an accuracy of $86\%$ with sensitivity and specificity of $96\%$ and $84\%$ respectively. Note that, the accuracy and specificity scores are similar to the state-of-the-art reported in \cite{reasat2017detection}, while the sensitivity is superior by $11\%$. We observe similar results in other channel configuration and disease predictions. 

In summary, the generative modeling based two-stage architecture produces highly accurate models in limited channel settings, even with datasets that are distributed non-uniformly. This motivates the use of Seq2Seq models in clinical modeling pipelines to both exploit unsupervised data and build more meaningful metrics, that can in turn lead to better supervisory models.


\section{Acknowledgments}
We thank Dr. Girish Narayan (cardiologist) for lending his expertise in the experimental design. We also thank Rushil Anirudh for helping us with the data preparation. This work was performed under the auspices of the U.S. Dept. of Energy by Lawrence Livermore National Laboratory under Contract DE-AC52-07NA27344.
%

\bibliographystyle{IEEEtran}
\bibliography{refs}

\end{document}